\documentclass[10pt,twocolumn,letterpaper]{article}

\usepackage[pagenumbers]{wacv} 

\usepackage{graphicx}
\usepackage{amsmath}
\usepackage{amssymb}
\usepackage{booktabs}
\usepackage{algorithm}
\usepackage{algpseudocode}
\usepackage{amsmath,amsfonts}
\usepackage{textcomp}
\usepackage{xcolor}
\usepackage{xspace}
\usepackage{xurl}
\usepackage{multirow}
\usepackage{ulem}
\usepackage[pagebackref,breaklinks,colorlinks]{hyperref}

\newcommand{\attackName}{AdvHeat\xspace}
\newcommand{\mypara}[1]{\noindent\textbf{#1.}\xspace}

\usepackage[capitalize]{cleveref}
\crefname{section}{Sec.}{Secs.}
\Crefname{section}{Section}{Sections}
\Crefname{table}{Table}{Tables}
\crefname{table}{Tab.}{Tabs.}

\begin{document}

\title{Turn Fake into Real:\\Adversarial Head Turn Attacks Against Deepfake Detection}
\author{
  Weijie Wang\textsuperscript{1,2},
  Zhengyu Zhao\textsuperscript{3}\thanks{Corresponding author},
  Nicu Sebe\textsuperscript{1},
  Bruno Lepri\textsuperscript{2},\\
  {\tt\small \{weijie.wang,niculae.sebe\}@unitn.it, zhengyu.zhao@xjtu.edu.cn, lepri@fbk.eu}\\
  \textsuperscript{1}University of Trento, Italy \quad
  \textsuperscript{2}Fondazione Bruno Kessler, Italy
  \quad
  \textsuperscript{3}Xi'an Jiaotong University, China 
  \\
}

\maketitle

\begin{abstract}
Malicious use of deepfakes leads to serious public concerns and reduces people's trust in digital media.
Although effective deepfake detectors have been proposed, they are substantially vulnerable to adversarial attacks.
To evaluate the detector's robustness, recent studies have explored various attacks.
However, all existing attacks are limited to 2D image perturbations, which are hard to translate into real-world facial changes.
In this paper, we propose \textbf{adv}ersarial \textbf{hea}d \textbf{t}urn (\attackName), the first attempt at 3D adversarial face views against deepfake detectors, based on face view synthesis from a single-view fake image.
Extensive experiments validate the vulnerability of various detectors to \attackName in realistic, black-box scenarios.
For example, \attackName based on a simple random search yields a high attack success rate of 96.8\% with 360 searching steps.
When additional query access is allowed, we can further reduce the step budget to 50.
Additional analyses demonstrate that \attackName is better than conventional attacks on both the cross-detector transferability and robustness to defenses.
The adversarial images generated by \attackName are also shown to have natural looks.
Our code, including that for generating a multi-view dataset consisting of 360 synthetic views for each of 1000 IDs from FaceForensics++, is available at \url{https://github.com/twowwj/AdvHeaT}.

\end{abstract}

\section{Introduction}

\begin{figure*}[!t]
\centering
\includegraphics[width=0.9\textwidth]{ 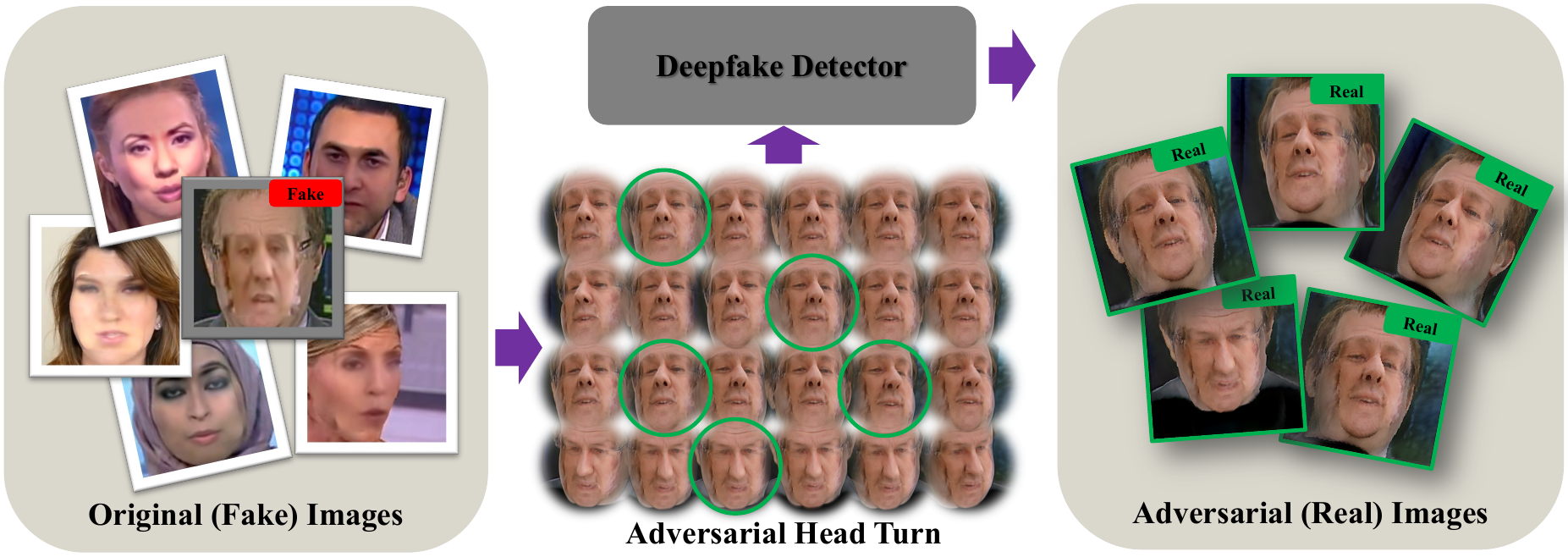}
  \caption{An illustration of our \textbf{adv}ersarial \textbf{hea}d \textbf{t}urn (\attackName) attack, which fools a deepfake detector to misclassify a fake image (left) as real (right) by synthesizing novel adversarial views (middle). Note that here we deliberately choose the fake images with obvious artifacts to show that even such fake images can become adversarial real images using our \attackName.}
  \label{fig:overview}
\end{figure*}

With the great power of deep learning, especially generative modeling, it becomes increasingly easy to automatically manipulate visual content.
A popular example of such manipulation is deepfakes~\cite{verdoliva2020media}, which replace the face of one person with that of another person. 
The misuse of deepfakes leads to a loss of trust in digital content and even assists the deliberate dissemination of disinformation~\cite{masood2022deepfakes}.

To combat such misuse, recent studies have been deploying significant efforts to develop forensic techniques for detecting deepfakes~\cite{rossler2019faceforensics++,verdoliva2020media}.
The state-of-the-art detection approaches are based on training a deep neural network (DNN). 
On the one hand, these approaches can effectively distinguish real from fake images.
On the other hand, the use of DNNs makes them substantially vulnerable to adversarial attacks~\cite{szegedy2014intriguing,goodfellow2015explaining}, which slightly perturb fake images to bypass the detection.

Understanding the vulnerability of deepfake detectors to adversarial attacks is necessary to achieve robust detection.
Most related work follows conventional adversarial attack literature to focus on face-agnostic image perturbations~\cite{carlini2020evading,gandhi2020adversarial,hussain2021adversarial,liao2021imperceptible,neekhara2021adversarial,jia2022exploring}, and a few studies have explored face attribute manipulations in the generator's latent space~\cite{carlini2020evading,li2021exploring}.
However, both two aspects of studies can hardly represent adversarial deepfakes in real-world facial applications, such as liveness tests.\footnote{\url{https://www.theverge.com/2022/5/18/23092964/deepfake-attack-facial-recognition-liveness-test-banks-sensity-report}}
In particular, although face attributes, e.g., skin color, hair color, and freckles, can be transformed into physical face changes, these changes are certainly hard to achieve in real-time.
In addition, face attribute manipulations need to use a generator to generate new fake images from scratch rather than modifying any existing fake image. 

In this paper, we propose \textbf{adv}ersarial \textbf{hea}d \textbf{t}urn (\attackName), a new type of adversarial deepfakes that is achieved by synthesizing 3D novel face views from a single-view fake face.
Figure~\ref{fig:overview} gives a high-level illustration of our \attackName.
\attackName is inspired by the fact that existing deepfake datasets typically concentrate on the front view of faces~\cite{rossler2019faceforensics++}. 
Technically, \attackName goes beyond the 2D image transformations in existing attacks.
More importantly, it represents a real-world threat in the context of real-time deepfakes in video calls.\footnote{\url{https://metaphysic.ai/to-uncover-a-deepfake-video-call-ask-the-caller-to-turn-sideways/}}
In this case, \attackName can synthesize adversarial fake images with different face views to bypass the deepfake detector.

One recent study~\cite{dong2022viewfool} also explores adversarial views, but there are two fundamental differences with respect to our work.
First, they focus on object classification while we focus on deepfake detection.
Second, they assume access to multiple original views (e.g., 100 in their paper) and leverage all these available images for novel view synthesis.
Differently, we assume that only a single-view image can be used for synthesis.
This assumption is realistic in deepfake detection since the deepfake images/videos available online are mostly with the fixed (front) view.  
Due to these two fundamental differences, the techniques used for viewpoint modeling and attack optimization are also different.

We conduct comprehensive experiments to evaluate the vulnerability of existing deepfake detectors to our new \attackName attack.
Since gradient calculation is known to be intractable for 3D synthesis~\cite{dong2022viewfool}, we focus our evaluation on black-box attacks, which only require the forward pass of the synthesis. 
Black-box attacks are also generally more challenging than white-box attacks and lead to more practical implications~\cite{carlini2017towards,papernot2017practical}.
Our results show that even a naive random search-based attack can be effective.
This implies the critical vulnerability of deepfake detectors to face view changes.
Moreover, in the scenario where the confidence scores are available through queries, the attack efficiency can be largely boosted.

Our main contributions are summarized as follows:
    \begin{itemize}
        \item We propose adversarial head turn (\attackName), a novel adversarial attack against deepfake detectors based on 3D face view synthesis.
        Specifically, we design two variants of \attackName, with one based on a simple random search and the other one that targets efficiency with additional query access to the target model. 
        \item We comprehensively evaluate \attackName against 10 deepfake detectors that are trained on 3 different levels of data quality.
        We find that all these detectors are largely vulnerable to \attackName in black-box scenarios, and the attack performance on different detectors may be related to their data quality.
        \item We conduct additional analyses to demonstrate the superiority of \attackName over conventional attacks on both the cross-detector transferability and robustness to defenses.
        We also show the natural looks of the adversarial images generated by \attackName.
    \end{itemize}

\begin{figure*}[!t]
\centering   \includegraphics[width=1\textwidth]{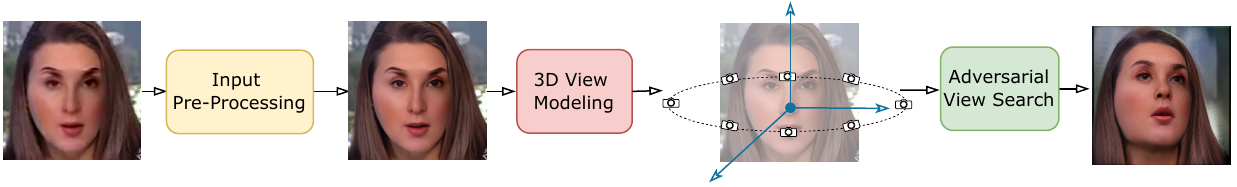}
  \caption{High-level illustration of our \attackName attack consisting of three modules: (i) input pre-processing, (ii) 3D view synthesis, and (iii) adversarial view search.
  }
  \label{fig:framework}
\end{figure*}

\section{Related Work}
\subsection{3D View Synthesis}
Neural Radiance Fields (NeRFs)~\cite{eslami2018neural, lombardi2019neural, martin2021nerf} learn implicit 3D representations by a neural network and then use it to synthesize novel views given 2D images with multiple views. The training of NeRFs only involves optimizing the parameters of the neural network using a collection of input images and their corresponding camera poses. Therefore, NeRFs have been widely used in different fields, such as 3D modeling \cite{wang2021neus} and human face reconstruction~\cite{gafni2021dynamic}. 

Recent work \cite{guo2021ad,schwarz2020graf} combines NeRFs with Generative Adversarial Networks (GANs) to design 3D-aware generators. It aims to control the pose (viewpoint) of the target object within the synthesized results and enhance the view consistency among multiple generated images. This is challenging for 2D generative models~\cite{karras2019style}. Using voxel-based methods \cite{lunz2020inverse} is a feasible approach compared to 2D generative models. However, it is difficult for high-resolution tasks and complex scenarios due to the high memory requirements. 
To address this limitation, a 3D-aware generator, Eg3D~\cite{chan2022efficient}, has been proposed based on the simple, 2D CNN-based  StyleGAN-v2~\cite{karras2020analyzing} rather than the inefficient 3D convolutions. 
In this work, we adopt Eg3D to create novel views of a single frontal-view face image.

\subsection{Deepfakes and Detection}
Deepfakes~\cite{korshunov2018deepfakes} is the technique of replacing the facial identity or expression of a source object with that of a target object through the use of computer vision or computer graphics approaches, which mainly includes four tasks~\cite{tolosana2020deepfakes, karras2019style, 8295149,liu2019stgan}: (i) face synthesis, (ii) face swap, (iii) face attributes editing, and (iv) face expression reenactment.
Most of these above tasks aim to generate realistic deepfake
with the help of 3D models \cite{thies2016face2face}, AutoEncoders \cite{tewari2017mofa}, or Generative Adversarial Networks \cite{karras2019style}.

To combat deepfakes, existing studies often rely on detecting low-level visual artifacts, such as pixel-level artifacts~\cite{rossler2019faceforensics++, bappy2019hybrid}, texture difference~\cite{li2018exposing}, and blending ghost~\cite{li2020face}. However, such low-level artifacts become hard to be detected under image degradation. Thus, recent methods~\cite{dong2022protecting,yang2018exposing} turn to utilize more semantically meaningful cues, e.g., inconsistency of head and face posture direction. 
In this work, we select 10 representative deepfake detectors to evaluate the effectiveness of our new attack.

\subsection{Adversarial Attacks on Deepfake Detectors}
Although deep learning models have achieved great success in various tasks, they are known to be vulnerable to adversarial attacks~\cite{szegedy2014intriguing,goodfellow2015explaining}.
Adversarial attacks have been extensively studied over the past decade.
In the ideal, white-box scenario, an attacker has full access to the model details and can optimize perturbations based on the gradient information~\cite{goodfellow2015explaining,carlini2017towards}.
In the more realistic, black-box scenario, either query-based~\cite{chen2017zoo,brendel2018decision,ilyas2018black,guo2019simple,chen2020hopskipjumpattack} or transfer-based attacks~\cite{liu2017delving,tramer2017space,dong2018boosting,zhao2021success,zhao2022towards} are developed.
Specifically, query-based attacks are optimized through querying the target model for the output scores/decisions, while transfer-based attacks are optimized on a (white-box) surrogate model and then directly used to fool the target model.

Since deepfake detectors are essentially machine learning models, they are inevitably vulnerable to adversarial attacks. 
Recently, evaluating the vulnerability of deepfake detectors to adversarial attacks has attracted increasing attention~\cite{carlini2020evading,gandhi2020adversarial,hussain2021adversarial,li2021exploring,liao2021imperceptible,neekhara2021adversarial,jia2022exploring}.
Regarding the attack vector, existing work mainly focuses on conventional, $L_p$-restricted perturbations but also explores unrestricted perturbations that are obtained by disrupting the latent space of a generative mode~\cite{carlini2020evading,li2021exploring} or the frequency space~\cite{jia2022exploring}.
Note that latent space-based attacks generate images from scratch rather than modifying existing images. 
Regarding the attacker's knowledge, existing work has explored both the white-box and black-box scenarios.

In this work, we evaluate the vulnerability of deepfake detectors against a novel type of attack that is based on 3D view synthesis.
We focus on the relatively realistic, black-box scenario, involving both the score-based and decision-based attacks. 

\section{Adversarial Head Turn}
In this section, we present adversarial head turn (\attackName), our novel attack to fool the deepfake detector. 
\attackName adopts 3D modeling techniques to model 3D face views from a single-view image and then searches for specific view parameters that cause adversarial effects.
Figure~\ref{fig:framework} illustrates the whole process of \attackName, which consists of three steps: (i) input pre-processing, (ii) 3D view modeling from a single-view image, and (iii) adversarial view search.
Before detailing these three steps, we first briefly formulate the problem.

\mypara{Problem Formulation}
The attack problem can be formulated as follows.
Given a deepfake detector $f(\boldsymbol{x}):\boldsymbol{x}\in\mathcal{X}\to y\in\mathcal{Y}=\{\textrm{real},\textrm{fake}\}$ that predicts a binary label $y$ for an original forgery image $\boldsymbol{x}$, an attacker generates the adversarial example $\boldsymbol{x}'$ with a wrong prediction $y'$. More precisely, we search for specific camera parameters such that they can be used to synthesize an image $\boldsymbol{x}'$ with an adversarial view.
Following the literature~\cite{carlini2020evading,gandhi2020adversarial,hussain2021adversarial,li2021exploring,liao2021imperceptible,neekhara2021adversarial,jia2022exploring}, we focus on misclassification from fake to real, i.e., $y=\textrm{fake}$ and $y'=\textrm{real}$. 

\subsection{Input Pre-Processing}
The original images are sampled from doctored videos, which are usually collected from the web and have a suboptimal resolution. Most of the training data for the deepfake detector comes from low-quality, low-resolution forged data \cite{li2018exposing}. 
In addition, the subsequent 3D view synthesis may also introduce blurry effects. Thus, to enhance the quality of the original (fake) images, we adopt GFP-GAN-v3~\cite{wang2021towards}, which is originally developed for blind face restoration.
Specifically, we apply GFP-GAN-v3 to obtain a $2\times$ super-resolution version of an input image. 
To the best of our knowledge, so far there is no work that applies super-resolution to fool deepfake detectors. 

\subsection{3D View Synthesis from a Single View}
\label{sec:3d}

\mypara{Project the original image}
Current research on 3D view synthesis mainly focuses on directly generating multiple views from the latent space.
However, when it comes to synthesis for out-of-domain images, such as face identities of deepfake, the synthesis quality may not be good.
Thus, projecting the out-of-domain images back into the in-domain space is necessary.
To this end, we use the off-the-shelf model PTI~\cite{roich2021pivotal}
to invert an original image into a feature vector in the latent space of StyleGAN.
This latent space provides the best editability to reconstruct the input and help obtain large angles of face generation with pose consistency~\cite{chan2022efficient}.
The optimization is defined as follows:
\begin{equation}
    w^*, n^* =\arg\min_{w,n}\mathcal{L}_{\textrm{sim}}(x, \mathbb{G}(w, n)) + \lambda\mathcal{L}_{n}(n),
\end{equation}
where $x$ is the original (fake) image, $w$ is its corresponding latent feature vector, and $n$ is additive noise used for regularization. 
Essentially, it finds the optimal $w*$ by directly maximizing the LPIPS similarity \cite{zhang2018unreasonable} between the original image $x$ and the reconstructed image $\mathbb{G}(w, n)$. 
According to~\cite{roich2022pivotal}, only using the above optimization may lead to a $w^*$ with significant distortions.
Therefore, we further refine $w^*$ by fine-tuning the generator using the following loss function:
\begin{equation}
        \mathcal{L}_{ft}=\mathcal{L}_{\textrm{sim1}}(x, \mathbb{G}(w^*, n)) + \lambda\mathcal{L}_{\textrm{sim2}}(x, \mathbb{G}(w^*, n)),
\end{equation}
where the LPIPS is used for $\mathcal{L}_{\textrm{sim1}}$ and the $L_2$ norm is used for $\mathcal{L}_{\textrm{sim2}}$.
Finally, the fine-tuned model can be used in the following step to synthesize multiple views of any given single-view image via the latent feature vector.

\begin{figure}[!t]
  \includegraphics[width=\columnwidth]{ 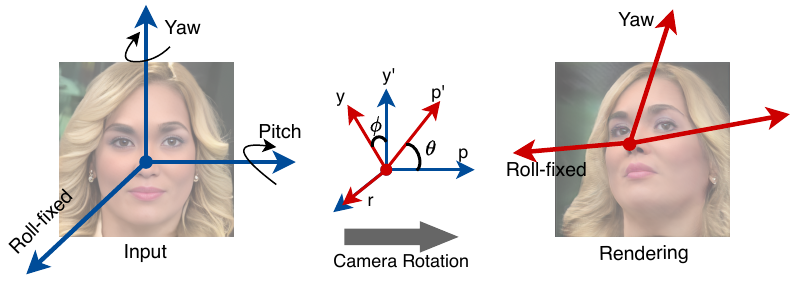}
  \caption{Description of viewpoint changes. The camera is fixed along the ``Roll'' axis and rotated by $\theta$ along the ``Pitch'' axis and $\phi$ along the ``Yaw'' axis.}
  \label{fig:viewpoint}
\end{figure}

\mypara{Synthesizing novel views}
We use the Eg3D~\cite{chan2022efficient} to generate images from multiple viewpoints with high quality and view consistency. 
Given the latent feature vector $w^*$ obtained from the last step, armed with the tri-plane representation, we use the trained 3D GAN framework of Eg3D
to produce multiple viewpoints by specifying the Euler angle parameters. 
In this paper, we fix the ``Roll'' axis and only ``turn the head'' along the ``Pitch'' axis $\phi$ and ``Yaw'' axis $\theta$. Given the rotation angles $(\phi, \theta)$,
the process can be described as follows:
\begin{equation}
\label{eq:turn}
\begin{aligned}
\left\{\begin{matrix}
    \phi_i = \frac{\pi}{2} + Y * Sin(2 \pi * i / K)  \\
    \theta_i = \frac{\pi}{2} + P * Cos(2 \pi * i / K)  
\end{matrix}\right.
\end{aligned}
\end{equation}
Here $K$ is the sampling frequency, which determines the total number of desired views.
$Y$ and $P$ are used to adjust the maximum allowed angles along the ``Pitch'' and ``Yaw'' axes.
Figure~\ref{fig:viewpoint} illustrates how the viewpoint changes along the two axes.

\subsection{Adversarial View Search}

Through the above steps, we are able to synthesize fake images with any novel face view.
Then, we need to find out which specific views could potentially cause the misclassification of the detector.
We focus on black-box attacks because they are more realistic than white-box attacks, and in addition, calculating gradients is computationally expensive in the context of 3D synthesis~\cite{dong2022viewfool}. 
To this end, we propose two methods to search for adversarial views. 
The first method is based on a random search, which only requires the final output label of the target detector.
The second method is based on gradient estimation, which requires additional query access to the model output confidence scores. 
We denote these two methods as \attackName-rand and \attackName-score, respectively.

\mypara{\attackName-rand}
Random search has been extensively used to find unrestricted adversarial examples~\cite{brown2018unrestricted}.
This is feasible because the attack space of unrestricted adversarial examples is commonly formulated with only a few parameters, in contrast to the high-dimensional space of additive pixel perturbations.
For example, existing work has applied random search to finding adversarial spatial image transformations~\cite{engstrom2019exploring} and color shifts~\cite{shamsabadi2020colorfool,zhao2023adversarial}. 
Here, in our case, the attack space is a one-dimensional space, represented by the index $i$ of the view in Equation~\ref{eq:turn}. 

\mypara{\attackName-score}
In practical attack scenarios, it is also realistic to have query access to the target model for obtaining the output confidence scores.
A common approach to such query-based attacks is zeroth-order optimization (ZOO)~\cite{ghadimi2013stochastic,chen2017zoo}.
ZOO computes an approximate gradient instead of the actual backpropagation on the target model in the white-box scenario.
Since an accurate gradient is usually unnecessary for successful adversarial attacks, ZOO is sufficient to achieve a high success rate~\cite{chen2017zoo}.

Following~\cite{chen2017zoo}, we formulate the gradient approximation based on the asymmetric difference quotient~\cite{lax2014calculus}:
\begin{equation}
   \label{eq:grad_estimate} 
   \hat{\boldsymbol{g}_k} : =  \frac{\partial l(\boldsymbol{x}_k)}{\partial \boldsymbol{x}} \approx 
\frac{l(\boldsymbol{x}_{k}) - l(\boldsymbol{x}_{k - h})}{h}, 
\end{equation}
where $l(\boldsymbol{x}_k)$ is the cross-entropy loss with respect to the image $\boldsymbol{x}_k$ at the index $k$.
We set the difference $h=1$.
Since our attack space is just one dimensional, we do not need a sophisticated optimization algorithm and additional dimension reduction as in~\cite{chen2017zoo} for high-dimensional, pixel-space adversarial attacks. 
Here, the step size $h$ controls the quality of the gradient estimate. Consequently, $h$ is a hyper-parameter in practice.

\begin{figure}[!t]
  \includegraphics[width=\columnwidth]{ 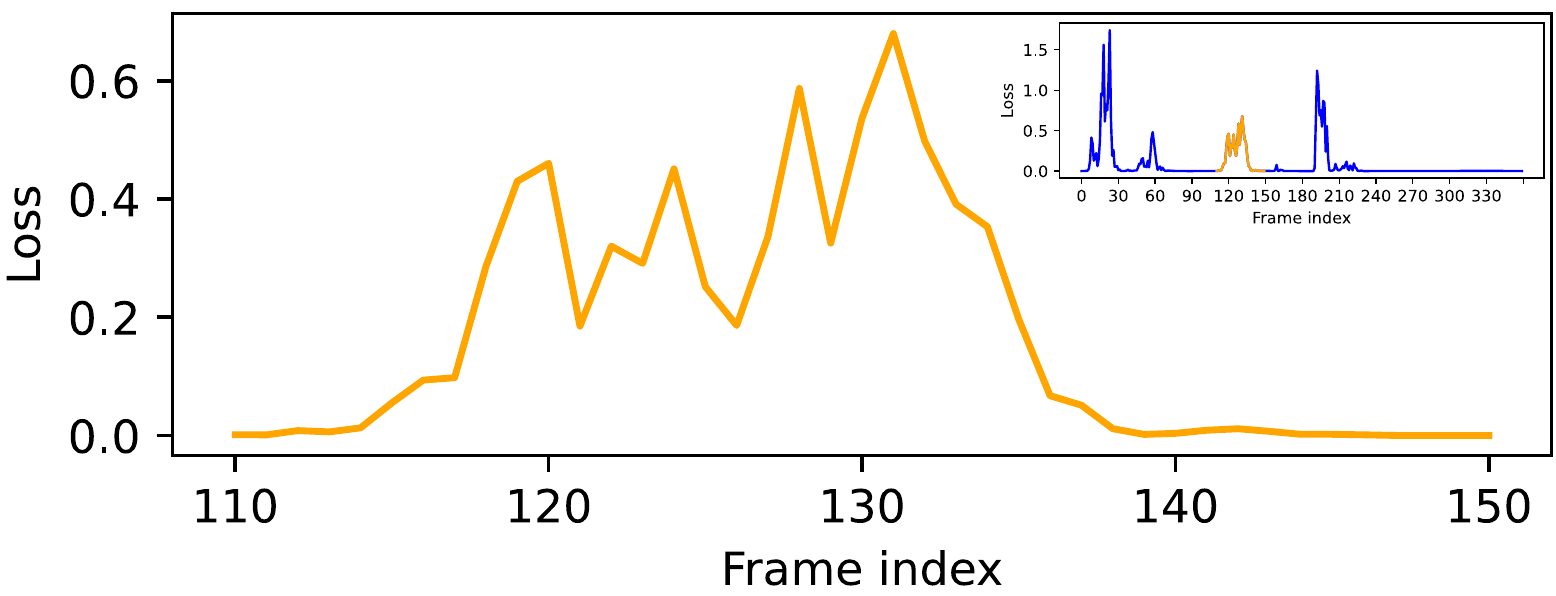}
  \caption{The loss distribution across different viewpoints for a random data sample.}
  \label{loss_index}
\end{figure}

However, our preliminary results suggest that directly using Equation \ref{eq:grad_estimate} does not yield good attack performance. 
Figure \ref{loss_index} explains this limitation by showing that the loss distribution exhibits significant fluctuations across different viewpoints.
This suggests that the gradients approximated based on the loss difference can hardly guide the search toward a global optimal.
To address this limitation, we only use the direction of the gradients for guidance.
The full procedure of \attackName-score is described in Algorithm~\ref{algorithm}.
Based on the approximated gradients, we apply iterative gradient descent to search for the adversarial view, starting from a randomly selected view from the total $K$ views.

\section{Experiments}
In this section, we evaluate the performance of our \attackName against various deepfake detectors, in terms of two important requirements of a successful attack: the success rate and the visual stealthiness. 
We also compare \attackName with other attack baselines in terms of cross-model transferability and robustness to defenses.
 
\subsection{Experimental Setups} 
\mypara{Dataset and Detectors} We use FaceForensics++~\cite{rossler2019faceforensics++}, a large-scale forensics dataset that consists of 1000 video sequences that have been manipulated with four different face manipulation methods: DeepFakes~\cite{li2020face}, Face2Face~\cite{2016Face2Face}, FaceSwap~\cite{2016Face2Face}, and NeuralTextrues~\cite{2019Deferred}.
Each video contains a trackable mostly-frontal face of identity without occlusions.
We randomly select one frame from each of these 1000 manipulated videos, resulting in an image dataset with a total of 1000 frontal face images.
Each image is cropped with a size of 317 $\times$ 317 using RetinaFace \cite{deng2020retinaface}.

We evaluate 10 representative deepfake detectors\footnote{We use the open-souce models provided in \url{https://github.com/wangjk666/PyDeepFakeDet}}: ResNet50 \cite{He2015DeepRL}, Xception \cite{Chollet2016XceptionDL}, EfficientNet \cite{tan2019efficientnet}, Meso4 \cite{Afchar2018MesoNetAC}, Meso4-Inc, \cite{Afchar2018MesoNetAC}, GramNet \cite{Liu2020GlobalTE}, F3Net \cite{Qian2020ThinkingIF}, MAT \cite{Zhao2021MultiattentionalDD}, ViT (transformer-based) \cite{Dosovitskiy2020AnII}, and M2TR (transformer-based) \cite{wang2022m2tr}.
For each detector architecture, we test 3 different quality of training images: Raw (i.e., without compression), high quality (HQ, with light compression), or low quality (LQ, with strong compression)~\cite{rossler2019faceforensics++}, resulting in 30 detectors in total.
All testing images are HQ images.

\begin{algorithm}[!t]
\caption{Attack Optimization for \attackName-score}
\label{algorithm}
\algrenewcommand\algorithmicrequire{\textbf{Input:}}
\algrenewcommand\algorithmicensure{\textbf{Output:}}
\algorithmicrequire{\\$\boldsymbol{x}$: Input fake image, 
$T$: maximum number of queries\\
$K$: sampling frequency, $\alpha$: step size\\
}
\algorithmicensure{ $\boldsymbol{x}'$: adversarial image}
\begin{algorithmic}[1]
\State Randomly select $k$ from $\{1,...,K\}$
\For {$t\leftarrow 0$ to $T-1$}
\State Calculate $\hat{\boldsymbol{g}_k}$ using Equation~\ref{eq:grad_estimate}
\State $k \leftarrow k+\alpha\cdot sgn(\hat{\boldsymbol{g}_k})$
\If{$\boldsymbol{x}_{k}$ is adversarial}
\State break
\EndIf
\EndFor
\State \Return $\boldsymbol{x}'\leftarrow\boldsymbol{x}_{k}$ 
\end{algorithmic}
\end{algorithm}

\mypara{Attack Hyperparameters}
If not explicitly mentioned, we set the total number of desired views, i.e., $K$ in Equation~\ref{eq:turn}, as 360 for both \attackName-rand and \attackName-score.
We show that $K=360$ is sufficient to achieve a good attack performance in Figure~\ref{fig:perturbation}.
We set $\phi$ $\in [-\pi/12,+\pi/12]$ following~\cite{chan2022efficient} and $\theta$ $\in [-\pi/2,+\pi/2]$ with the side view of a face as the largest angle.
For adversarial view search with \attackName-score, we decrease the step size $\alpha$ from 10 to 3 with cosine annealing.
We early stop the optimization when the loss value oscillates between two indexes and then repeat the process until it finds the adversarial image or reaches the maximum number of queries $T$.
If not explicitly mentioned, for both \attackName-rand and \attackName-score, we adopt $T=360$ for a fair comparison.
All our experiments are conducted on an NVIDIA TITAN Xp GPU. 
For each identity, it takes about 9 mins for 3D view modeling to generate 360 viewpoints.

\begin{table}[!t]
\caption{Attack success rates (\%) of \attackName on different detectors. Results are reported for both \attackName-rand/-score.}
\label{attack_success_rate}
\resizebox{\columnwidth}{!}{
\begin{tabular}{l|ccc}
\toprule
Detector& Raw& HQ&LQ   \\ 
\midrule 
ResNet50~\cite{he2016deep}                          &    67.2/64.8     &   69.1/67.9 &   92.7/90.9 \\
Xception~\cite{Chollet2016XceptionDL}               &    5.3 /4.6      &   51.4/49.7 &   99.6/97.4 \\
EfficientNet~\cite{tan2019efficientnet}          &    47.4/44.0     &   86.8/84.2 &   95.6/93.8 \\
ViT~\cite{Dosovitskiy2020AnII}        &    33.0/31.9     &   80.6/79.4 &   99.3/97.1 \\
\midrule 
F3Net~\cite{Qian2020ThinkingIF}                     &    98.0/97.8     &   98.5/98.3 & 99.9/99.8   \\
Multi-Attention~\cite{Zhao2021MultiattentionalDD}   &    68.9/65.1     &   99.6/99.0 & 99.9/99.7   \\
GramNet~\cite{Liu2020GlobalTE}                      &    25.0/23.2     &   54.0/52.7 & 99.4/97.8   \\
M2TR~\cite{wang2022m2tr}                            &    95.7/92.9     &   96.6/96.0 &  98.6/96.9  \\
Meso4~\cite{Afchar2018MesoNetAC}                  &    21.5/19.6     &   65.8/63.6 & 89.0/87.4   \\
Meso4-Inc~\cite{Afchar2018MesoNetAC}     &    24.6/21.4     &   74.3/71.2 & 94.5/92.8   \\
\midrule 
Average     &    49.0 /46.5    &    78.1/76.2 &  96.8/95.3 \\
\bottomrule  
\end{tabular}}
\end{table}

\begin{figure}[!t]
\includegraphics[width=\columnwidth]{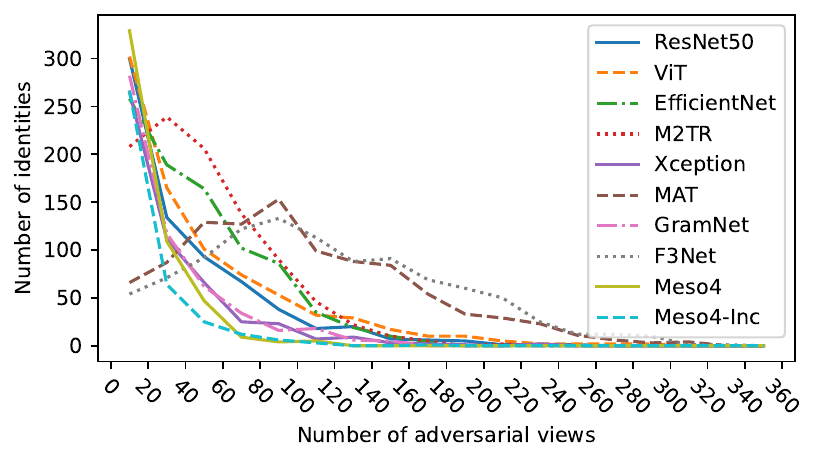}
  \caption{Histogram of the number of adversarial views over all 1000 identities in our dataset.}
    \label{fig:id_his}
\end{figure}

\subsection{Effectiveness of \attackName}
\mypara{Success rate} We first look at the attack success rates for different detectors and image quality.
The results are reported in Table~\ref{attack_success_rate}.
When comparing different image quality, we can see that it gets harder to attack the detector when the input image quality is increased (from LQ to Raw).
This may be because when the image quality is higher, there is more detailed information for the detector to make a more accurate prediction.

We can also see the attacks achieve different results on different detectors, especially for Raw. 
Interestingly, the detectors that are known to yield state-of-the-art detection accuracy, e.g., F3Net, M2TR, and MTA, are actually the most vulnerable ones against our attacks.
This is consistent with the commonly believed finding that adversarial robustness may be at odds with accuracy~\cite{tsipras2019robustness}.
Figure~\ref{fig:id_his} further supports this finding by showing that these three detectors lead to many more identities that have a large number of adversarial views than the other seven detectors.

\begin{figure}[!t]
\includegraphics[width=\columnwidth]{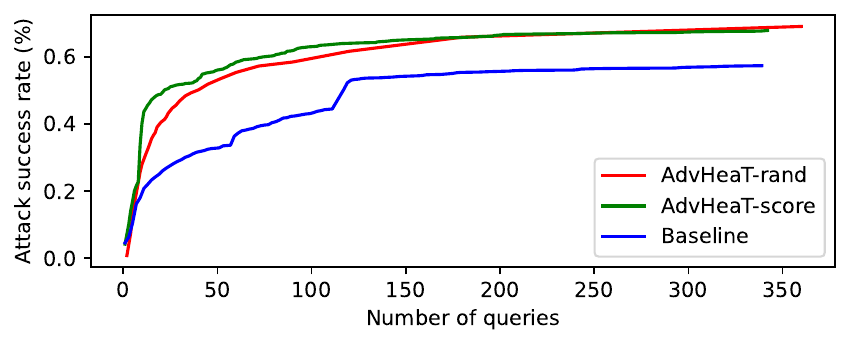}
  \caption{Attack success rates (\%) against the number of queries for \attackName-rand vs. \attackName-score on the ResNet50 detector.
  ``Baseline'' denotes the method that directly uses Equation~\ref{eq:grad_estimate}.}
  \label{fig:iter}
\end{figure}

\begin{figure}[!t]
\centering
\includegraphics[width=0.9\columnwidth]{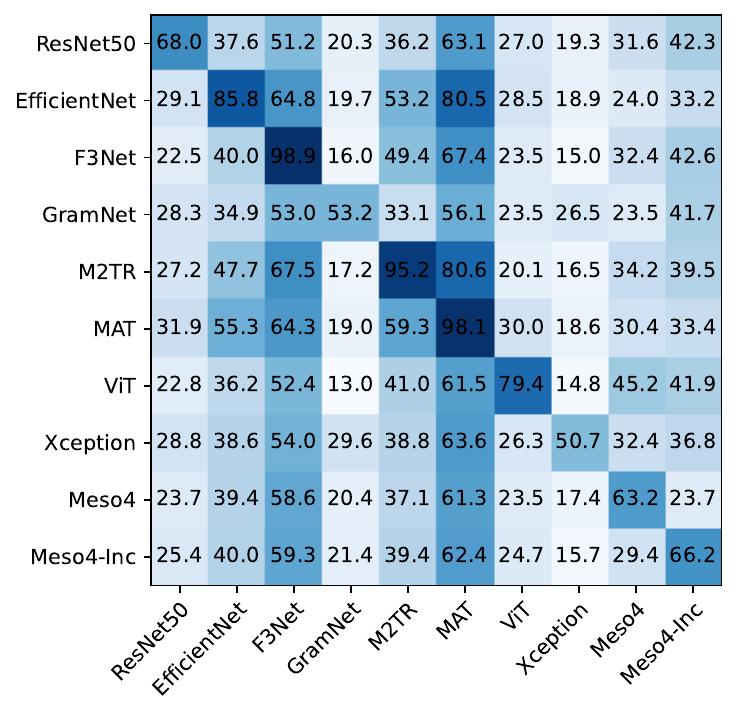}
  \caption{Cross-model transferability (\%) of \attackName.}
  \label{transfer_matrix}
\end{figure}

\begin{figure*}
\centering 
  \includegraphics[width=1\textwidth]{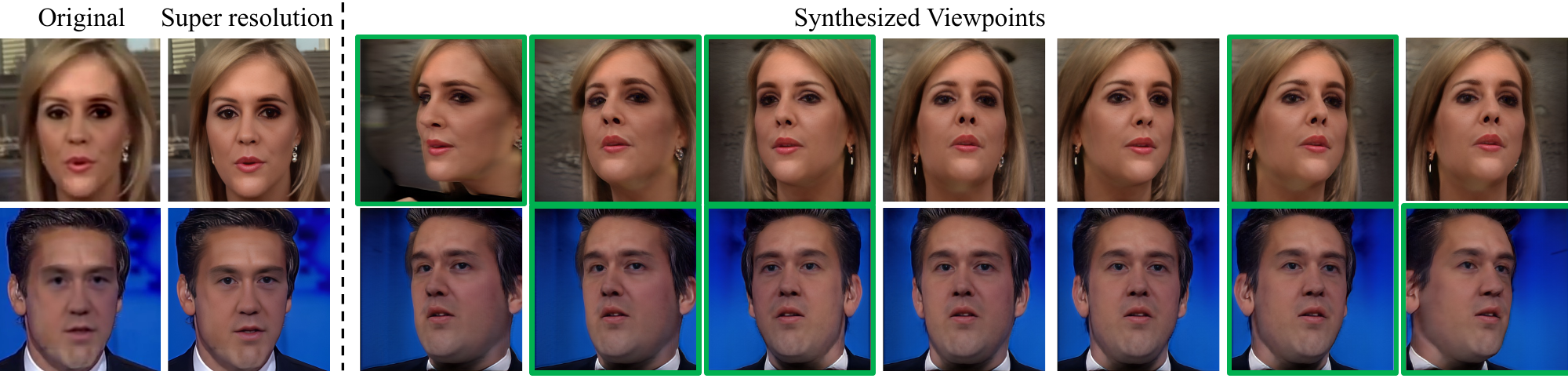}
  \caption{Visualizations of adversarial views (in \textcolor{green}{green} boxes) generated by our attack. ResNet50 detector is used. Additional examples can be found in Figure~\ref{fig:add_view}.}
  \label{visualization}
\end{figure*}

\begin{table*}[!t]
\centering
\caption{Comparing \attackName with other attacks in terms of cross-model transferability and robustness to defenses. ResNet50 detector is used as the original detector. Results are reported for Raw/HQ/LQ detectors.}
\label{tab:com}
\resizebox{\textwidth}{!}{
\begin{tabular}{l|cccc|cccc|c}
\toprule
Attack&\multicolumn{4}{c|}{Transferability}& \multicolumn{4}{c|}{Robustness}&\multirow{2}{*}{Average}\\ 
Method&M2TR&EfficientNet&ViT&F3Net&JPEG~\cite{dziugaite2016study}&R\&P~\cite{xu2017feature}&BDR~\cite{xu2017feature}&NRP~\cite{naseer2020self} &\\ 
\midrule
FGSM~\cite{goodfellow2015explaining}&\underline{29.2}/28.7/\underline{24.1}&0.2/24.9/0.0&\textbf{56.5}/\textbf{86.0}/26.1&27.0/25.2/0&99.6/54.5/18.3&35.3/54.7/19.9&\underline{99.7}/82.7/\underline{88.4}&39.3/15.8/0.2&39.0\\
BIM~\cite{kurakin2017adversarial}&6.4/\textbf{60.5}/17.9&\textbf{62.4}/\textbf{93.7}/0.4&\underline{31.2}/38.2/\underline{27.4}&1.6/35.7/0&\underline{99.9}/\underline{99.6}/\underline{98.1}&\textbf{96.2}/\textbf{100}/\textbf{98.8}&99.6/\underline{84.4}/82.2&\textbf{99.4}/\underline{82.3}/15.6&\underline{59.7}\\
CW~\cite{carlini2017towards}&0/7.0/20.6&4.5/\underline{49.7}/\underline{5.5}&3.6/\underline{60.2}/23.4&3.7/40.4/0&\textbf{100}/76.0/23.2&51.8/70.6/30.3&\textbf{100/95.7}/68.4&57.6/62.8/\underline{29.3}&36.0 \\
\midrule
\attackName&\textbf{51.2}/\underline{47.1}/\textbf{31.6}&\underline{23.0}/48.7/\textbf{48.4}&4.3/34.6/\textbf{75.7}&\textbf{57.1}/\textbf{63.0}/0&99.0/99.3/\textbf{98.7}&\underline{74.7}/\underline{82.6}/\underline{94.2}&78.3/70.0/\textbf{97.8}&\underline{92.2}/\textbf{92.2}/\textbf{79.9}&\textbf{64.3}\\
\bottomrule 
\end{tabular}
}
\end{table*}

We further evaluate the success rates of our attacks against the query number.
We also include the sub-optimal method, i.e., Equation~\ref{eq:grad_estimate}, that is guided directly by the approximated gradients rather than only their direction as in \attackName-rand. 
As shown in Figure~\ref{fig:iter}, in general, all methods achieve better results as more queries are used.
Specifically, \attackName-rand serves as an upper bound when enough queries are given to explore all the 360 candidate views.
However, \attackName-score is more efficient by leveraging additional gradient information, making it more favorable when only a limited number of queries are allowed.

\mypara{Cross-model transferability} 
In realistic scenarios, it may be difficult to access the output probability of the target model or even query it hundreds of times~\cite{debenedetti2023evading}.
Therefore, the transferability of attacks across different models is important~\cite{zhao2022towards}.
Here we test the transferability of our attack with one detector as the source model and another as the target model.
For each identity, we select the view that yields the highest logit for the ``real'' class from the 360 views.
Note that this selected view may not always be sufficient to change the prediction of the source detector.

Figure \ref{transfer_matrix} reports the results. 
As can be seen, in general, the transferability of our attack is high, with an average success rate of 36.7\% (excluding the diagonal values).
When comparing different models, we can draw a similar conclusion that adversarial robustness may be at odds with accuracy. 
For example, the average attack success rate against MAT is 69.5\% while that against Xception is only 21.3\%.

\mypara{Visual stealthiness}
Figure \ref{visualization} visualizes the novel views, including adversarial views found by our attack.
It can be seen that those adversarial views spread over different viewpoints.
Since our attack modifies the original image in a semantic space represented by the head turn, the resulting adversarial images still enjoy natural looks.

\subsection{Comparisons with Other Attacks}
We further compare \attackName with well-known conventional attacks: FGSM~\cite{goodfellow2015explaining}, BIM~\cite{kurakin2017adversarial}, and CW~\cite{carlini2017towards}.
Specifically, FGSM and BIM are constrained by $L_{\infty}=8$, and BIM adopts 10 steps of gradient descent.
CW is constrained by $L_{2}$ with 50 steps and the balancing factor as 1.

\begin{figure}[!t]
\includegraphics[width=\columnwidth]{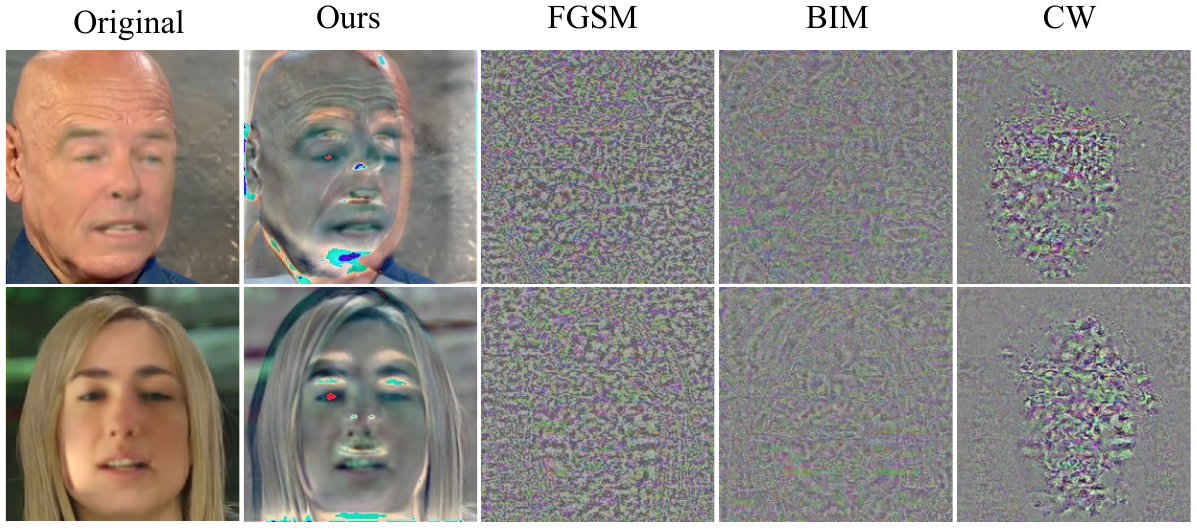}
  \caption{Visualizations of perturbations. Additional examples can be found in Figure~\ref{fig:add_per}.}
  \label{fig:perturbation}
\end{figure}

Table~\ref{tab:com} shows that on average, our \attackName achieves the best performance on both the transferability and robustness to defenses. 
Specifically, in most (16 out of 24) cases, it achieves the best or the second-best performance. 
Especially, this superiority is substantial for LQ detectors.
As shown in Figure \ref{fig:perturbation}, \attackName leads to perturbations that reflect obvious semantics of the identity rather than introducing semantically irrelevant artifacts.

\section{Ablation Studies}
In this section, we conduct ablation studies on the main components of \attackName, i.e., super-resolution and 3D view modeling, as well as important hyperparameters, i.e., the sampling frequency $K$ and the turning angles $\phi$ and $\theta$.

\mypara{Super-resolution and 3D view modeling}
Table~\ref{tab:ab} demonstrates the impact of these two main components on the attack performance.
As can be seen, using either of these two components leads to substantial attack success rates, and using both yields the best results.
This conclusion applies across all 3 different levels of image quality.

\begin{table}[!t]
\caption{The impact of the 3D view modeling and super-resolution (SR) on the attack success rates (\%) of \attackName-rand/-score. Results are averaged over the 10 detectors.}
\resizebox{\columnwidth}{!}{
\begin{tabular}{l|ccc}
\toprule
Method& Raw & HQ & LQ \\ 
\midrule 
Original& 0.5/0.4 &0.4/0.4 & 6.6/6.1 \\
~+3D    &17.5/17.2 &36.0/35.9  &  77.1/75.2    \\
~+SR    &21.1/20.8&43.8/42.9 & 49.7/47.8  \\
\midrule 
~+3D+SR (\attackName)&\textbf{49.0}/\textbf{46.5}&\textbf{78.1}/\textbf{76.2}   & \textbf{96.8/95.3}   \\
\bottomrule  
\end{tabular}
}
\label{tab:ab}
\end{table}

\begin{figure}[!t]
\includegraphics[width=\columnwidth]{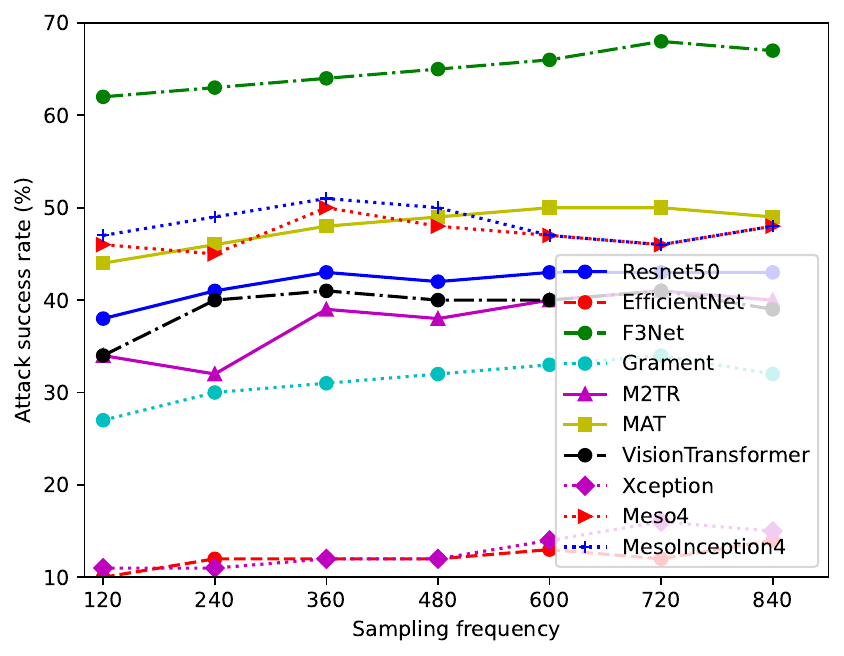}
  \caption{The impact of the sampling frequency $K$ on \attackName.}
  \label{sample_frequency}
  \vspace{-0.25cm}
\end{figure} 

\mypara{Sampling frequency $K$}
As described in Section~\ref{sec:3d}, for each identity, we should first synthesize $K$ candidate views, from which we would select the adversarial views.
Intuitively, a small $K$ limits the number of candidate views and consequently limits the number of potentially adversarial views. 
On the contrary, a large $K$ consumes a large number of computations due to the time-consuming nature of 3D synthesis.
Therefore, setting a proper $K$ is crucial for achieving a good trade-off between the attack performance and efficiency.
As can be seen from Figure~\ref{sample_frequency}, setting a very low value for $K$ would obviously limit the attack success but it is not always optimal to set a very high value.
To balance the attack performance and efficiency, we set $K=360$ in our main experiments.

\begin{figure}[!t]
\includegraphics[width=\columnwidth]{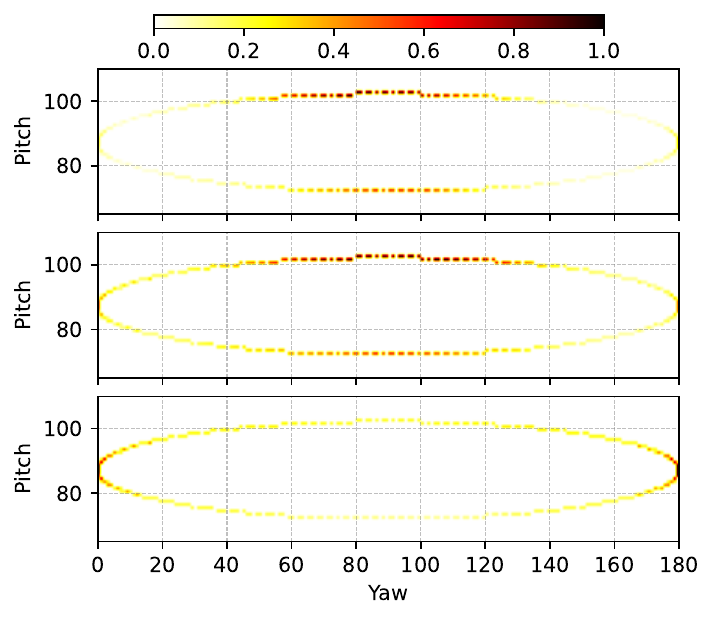}
\caption{The impact of turning angles $\phi$ and $\theta$ on \attackName for (top) Raw, (middle) HQ, and (bottom) LQ detectors. Results are averaged over the 10 detectors.}
\label{fig:axes}
\end{figure}

\begin{figure}[!t]
\includegraphics[width=\columnwidth]{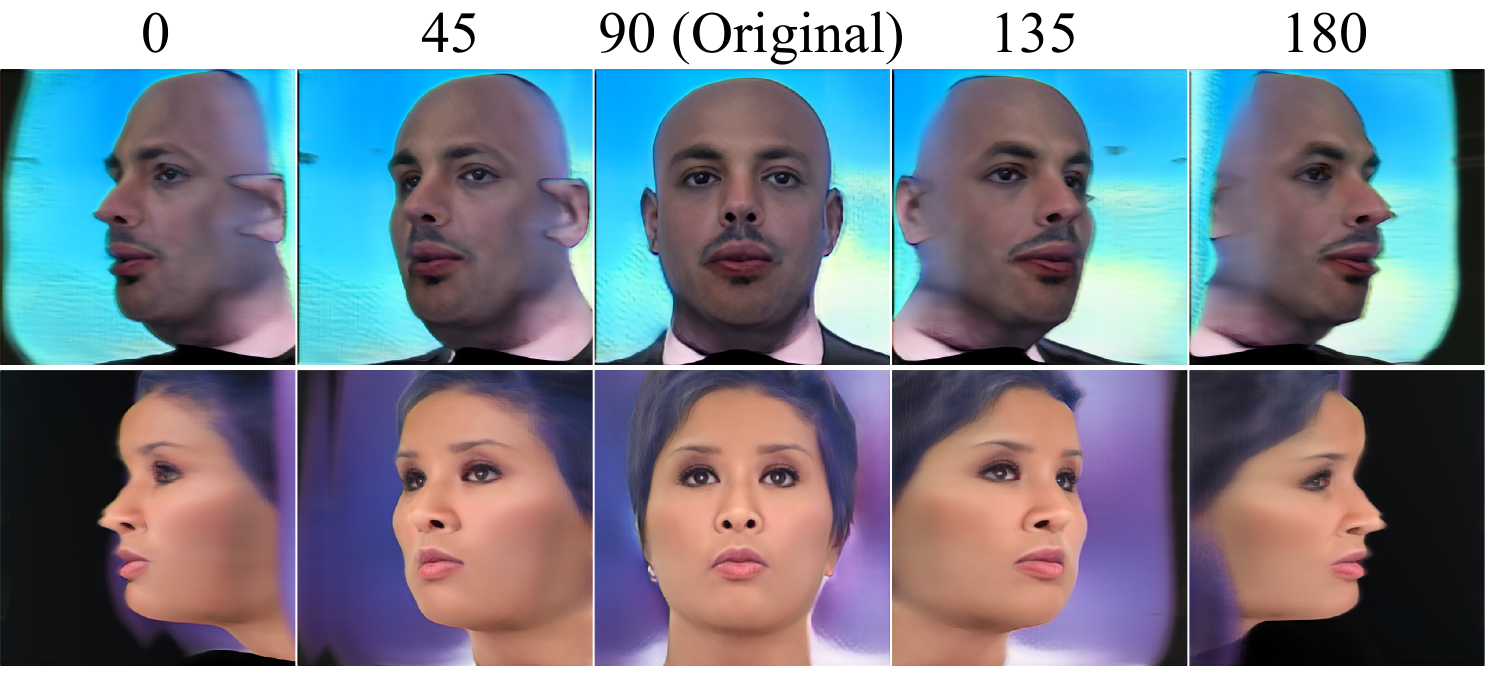}
\caption{Too large turning angles lead to low synthesis quality.}
\label{fig:exp}
\end{figure}

\mypara{Turning angles $\phi$ and $\theta$}
Figure~\ref{fig:axes} presents the attack success calculated at different turning angles $\phi$ and $\theta$. 
As can be seen, for the Raw and HQ detectors, large angles along the two axes result in very few adversarial views.
This makes sense because when the head is turned too much, the resulting image with low synthesis quality, as shown in Figure~\ref{fig:exp}, is hardly predicted to be real by the detector.
This finding also demonstrates that our setting with $\phi$ $\in [-\pi/12,+\pi/12]$ and $\theta$ $\in [-\pi/2,+\pi/2]$ is sufficient to guarantee the optimal attack success. 

However, somewhat surprisingly, for the LQ detector, most adversarial views are found at very large angles, especially along the ``Yaw'' axis.
This may be explained by the fact that the LQ detector has been trained on low-quality images. As a result, images with low synthesis quality at very large angles still have a high chance of misleading the detector.
We leave more detailed explorations about the relation between the quality of training data and the adversarial robustness for future work.

\section{Conclusion and Outlook}
In this paper, we propose \attackName, a novel approach to generating adversarial images against deepfake detectors.
\attackName is based on 3D view synthesis and uses two different methods to search for adversarial views. 
Specifically, \attackName-rand assumes no access to the target model and relies on the random search, while \attackName-score assumes query access and approximate gradients to update the attack.
Extensive experiments demonstrate the substantial vulnerability of various deepfake detectors to our \attackName in realistic, black-box attack scenarios.
The generated adversarial images are also shown to have natural looks. 

In the future, we would extend our exploration by designing effective defenses against our new \attackName attack.
It remains a possibility that our methodology may be misused by malicious actors to bypass (legitimate) deepfake detectors. 
However, we firmly believe that given our comprehensive presentation of the new attack surface with 3D view modeling, the community would be well motivated to improve the robustness of deepfake detectors.

{\small
\bibliographystyle{ieee_fullname}
\bibliography{egbib}
}

\clearpage
\newpage

\begin{figure*}[!t]
     \centering
     \begin{subfigure}[b]{\textwidth}
         \centering
\includegraphics[width=\textwidth]{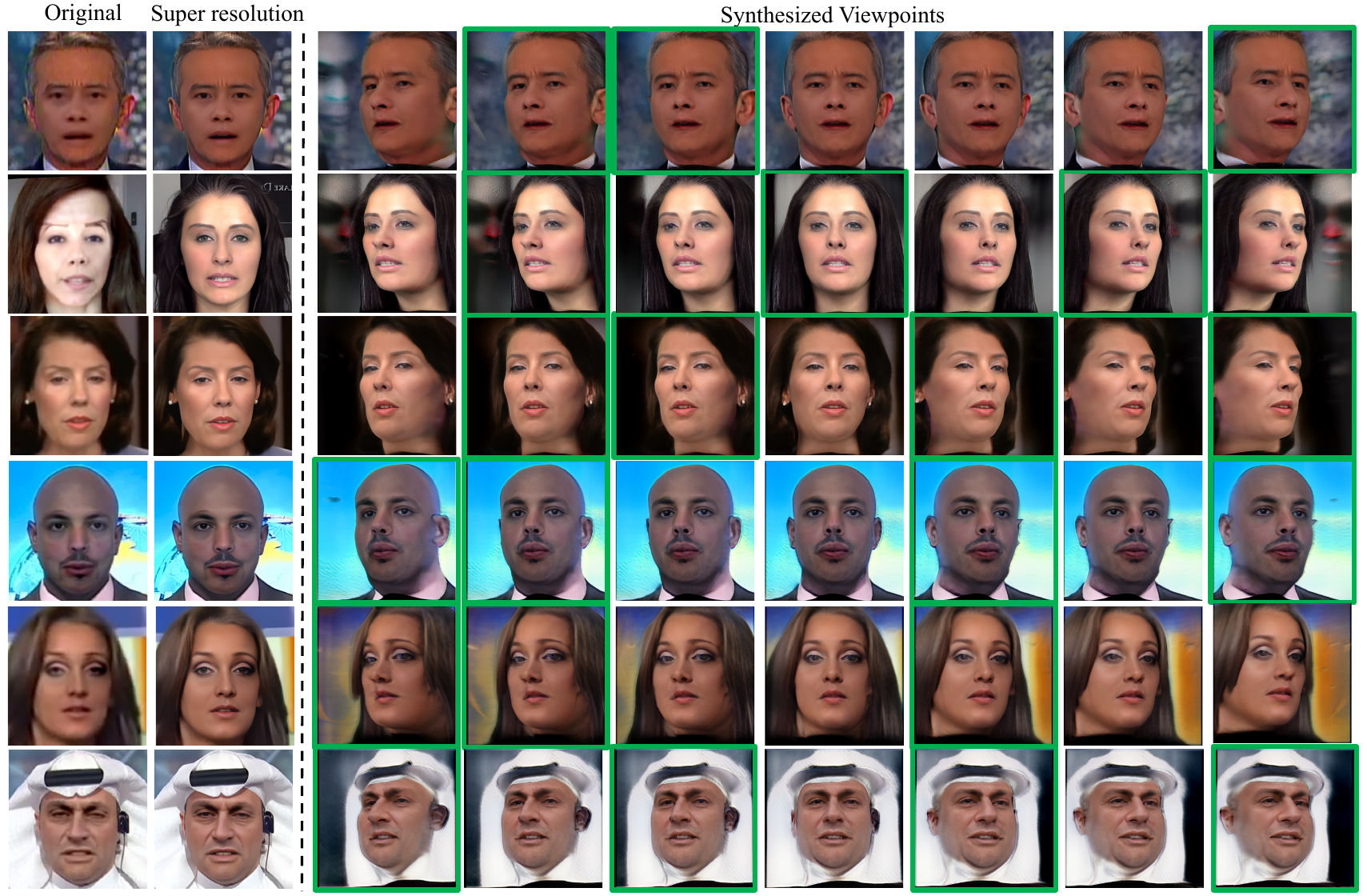}
         \label{sup:sup_vis_1}
     \vspace{-5mm}
     \end{subfigure}
     \hfill
     \begin{subfigure}[b]{\textwidth}
         \centering
\includegraphics[width=\textwidth]{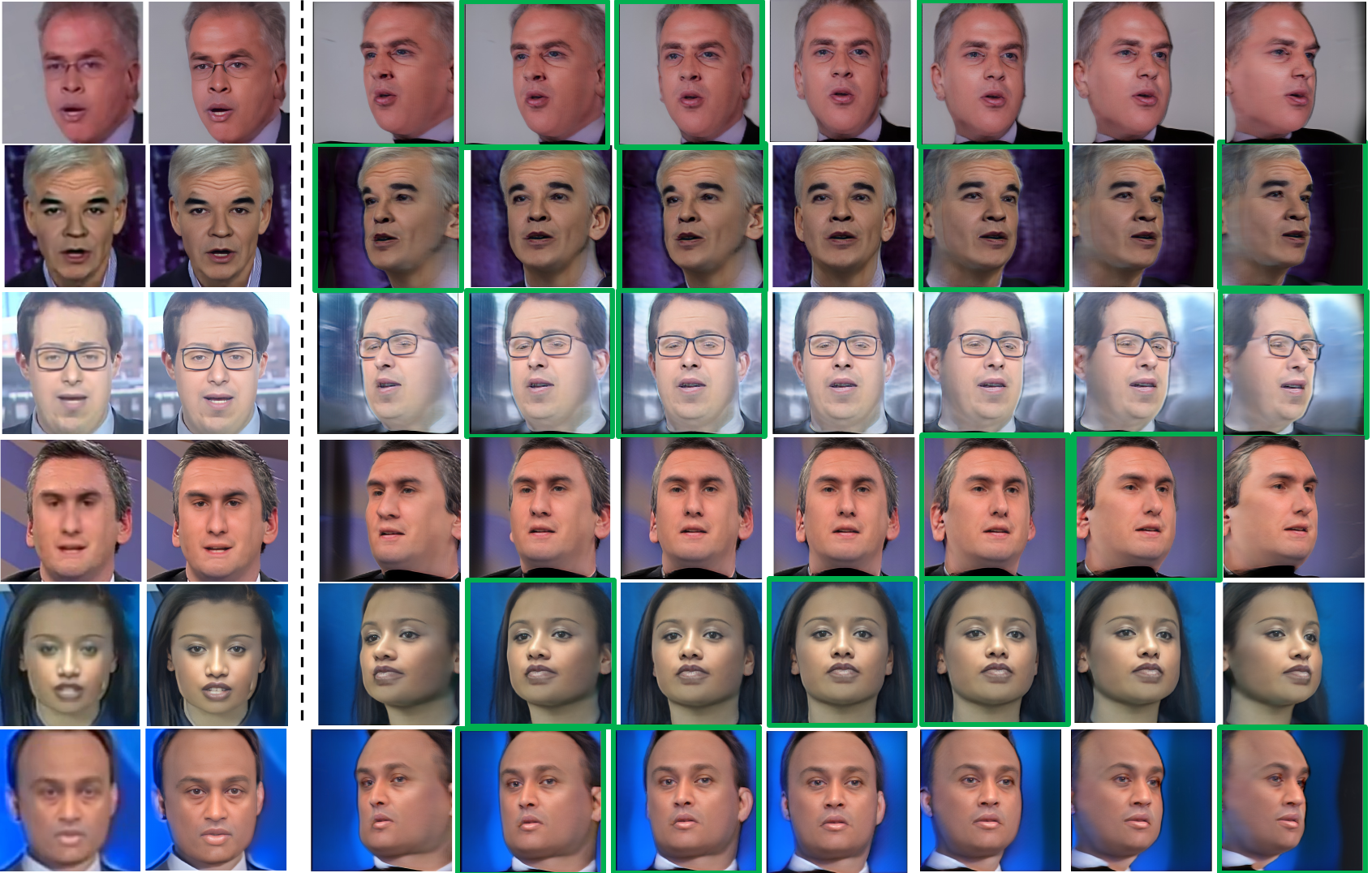}
          \label{sup:sup_vis_2}
     \end{subfigure}
     \vspace{-1cm}
     \caption{Additional examples of adversarial views (in \textcolor{green}{green} boxes) generated by our attack. ResNet50 detector is used.}
          \label{fig:add_view}
\end{figure*}

\clearpage
\newpage

\begin{figure*}[!t]
    \centering
    \begin{subfigure}{\columnwidth}
\includegraphics[width=\linewidth]{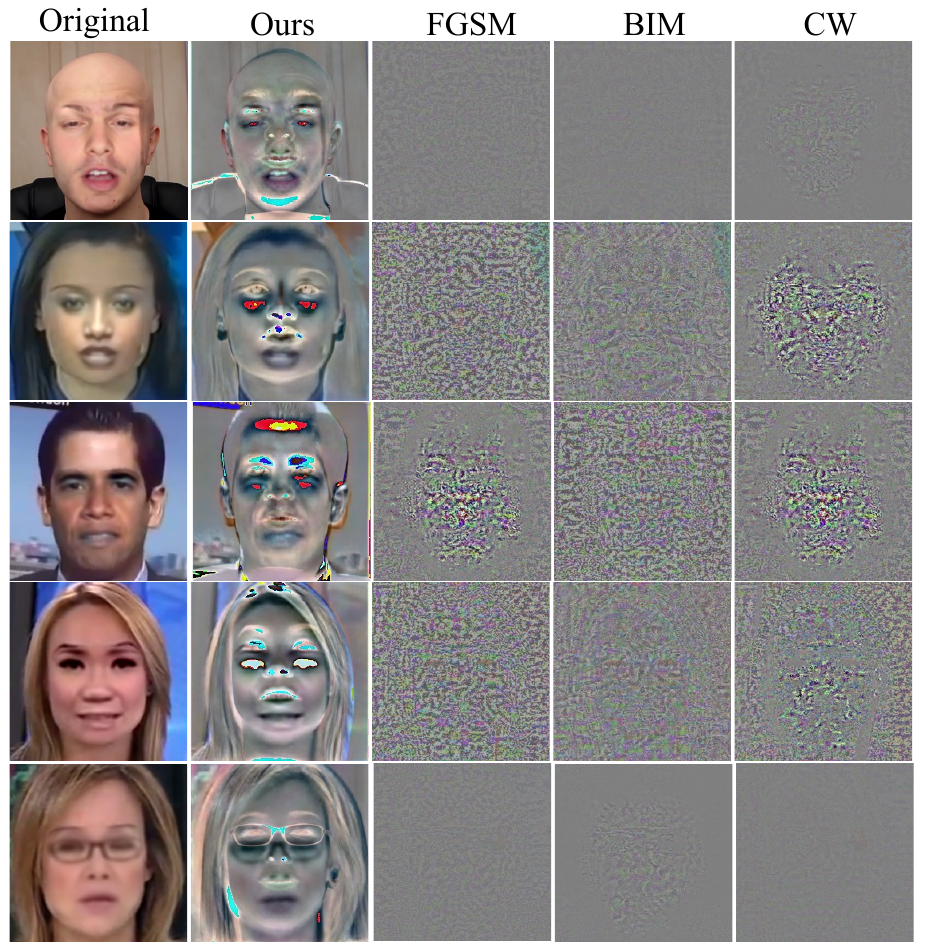}
\vspace{-1cm}
    \end{subfigure}
     \hfill
    \begin{subfigure}{\columnwidth}
\includegraphics[width=\linewidth]{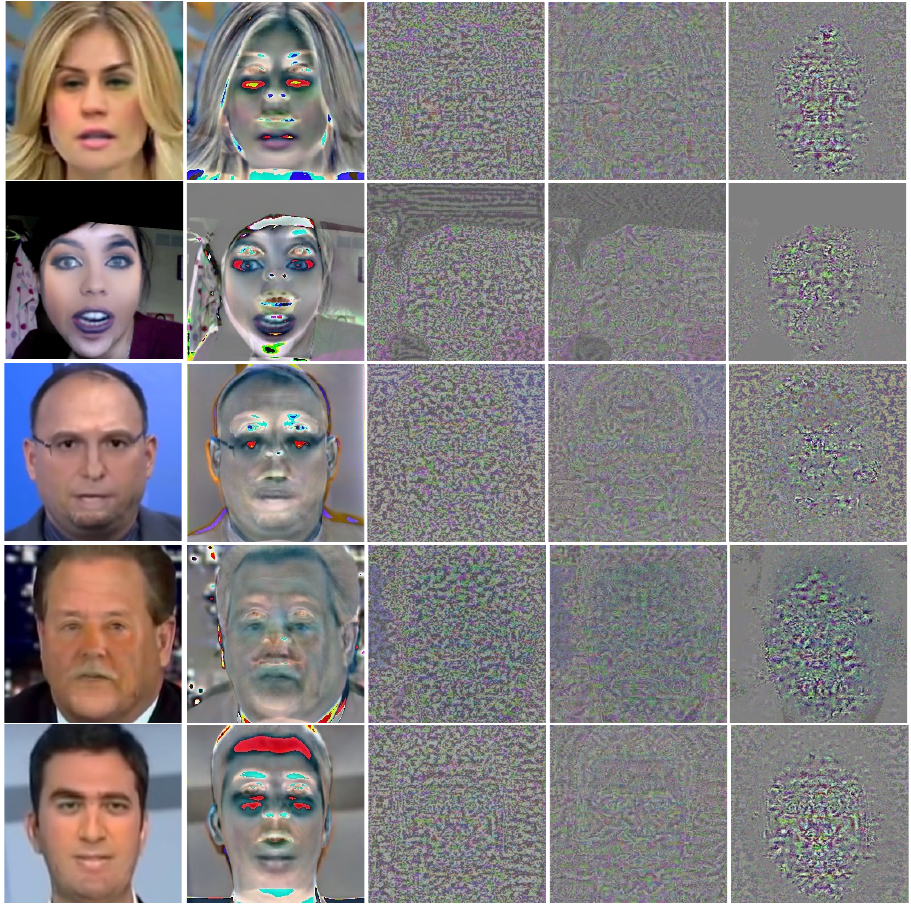}
    \end{subfigure}
    \begin{subfigure}{\columnwidth}
\includegraphics[width=\linewidth]{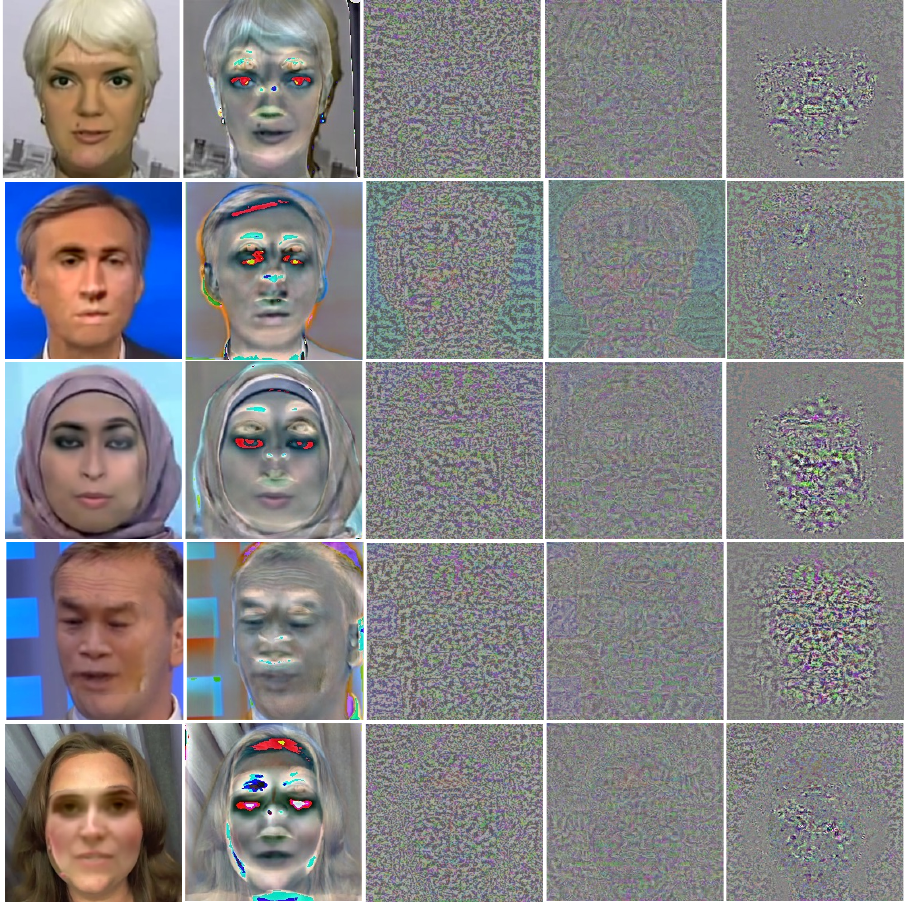}
\vspace{-1cm}
    \end{subfigure}
     \hfill
    \begin{subfigure}{\columnwidth}
\includegraphics[width=\linewidth]{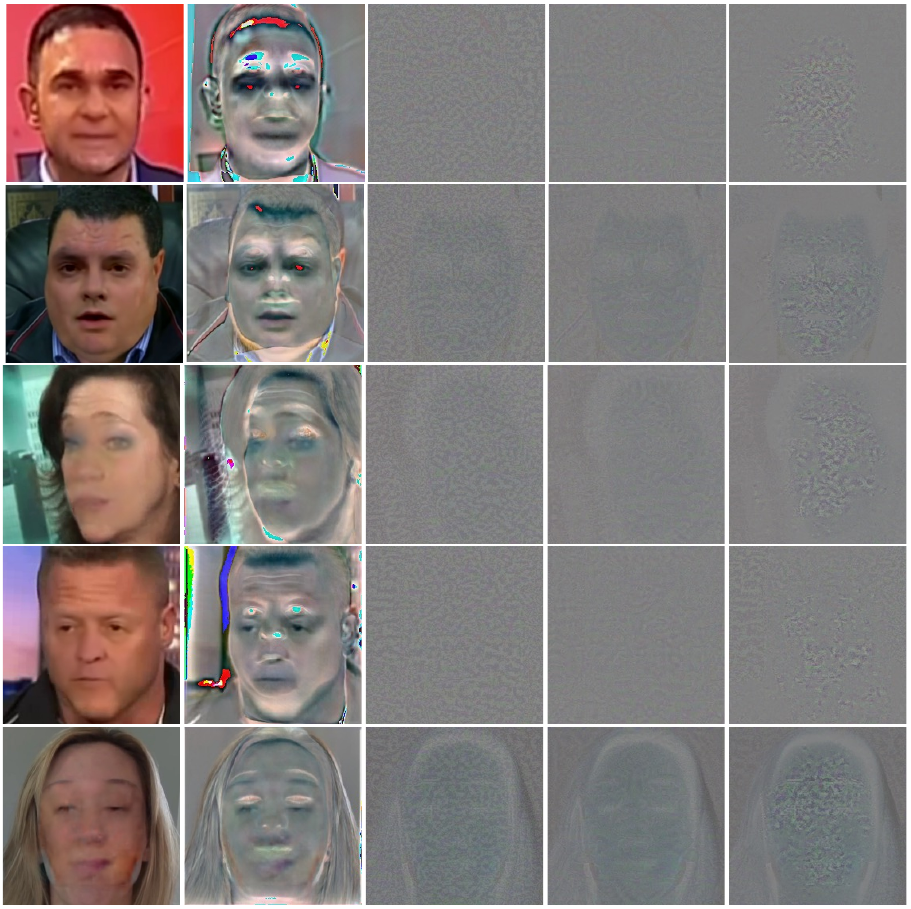}
    \end{subfigure}
    \vspace{2cm}
       \caption{Additional examples of perturbations.}
           \label{fig:add_per}
\end{figure*}
\end{document}